%% file: main.tex
\documentclass[10pt,twocolumn,letterpaper]{article}

\usepackage[pagenumbers]{iccv} 

\usepackage{times}
\usepackage{epsfig}
\usepackage{graphicx}
\usepackage{amsmath}
\usepackage{amssymb}
\usepackage{multirow}
\usepackage{bm}
\usepackage[ruled]{algorithm2e} 

\usepackage{xcolor,colortbl}
\definecolor{light}{gray}{0.95}

\input{preamble}

\definecolor{iccvblue}{rgb}{0.21,0.49,0.74}
\usepackage[pagebackref,breaklinks,colorlinks,allcolors=iccvblue]{hyperref}

\def\ourMethod{{TrajectoryCrafter}}

\newcommand*{\affaddr}[1]{#1} 
\newcommand*{\affmark}[1][*]{\textsuperscript{#1}}
\newcommand*{\email}[1]{\texttt{#1}}

\usepackage{lipsum}

\title{\ourMethod: Redirecting Camera Trajectory for Monocular Videos \\ via Diffusion Models}

\author{
Mark Yu \quad 
Wenbo Hu\affmark[\dag] \quad 
Jinbo Xing\affmark[] \quad 
Ying Shan\affmark[] \vspace{2mm}\\
\affaddr{\affmark[]ARC Lab, Tencent PCG} \quad \quad \quad
\affaddr{\affmark[]The Chinese University of Hong Kong} \\
\email{\url{https://TrajectoryCrafter.github.io}}
}

\begin{document}

\twocolumn[{%
\renewcommand\twocolumn[1][]{#1}%
\maketitle
\vspace{-1em}
\includegraphics[width=1.0\linewidth]{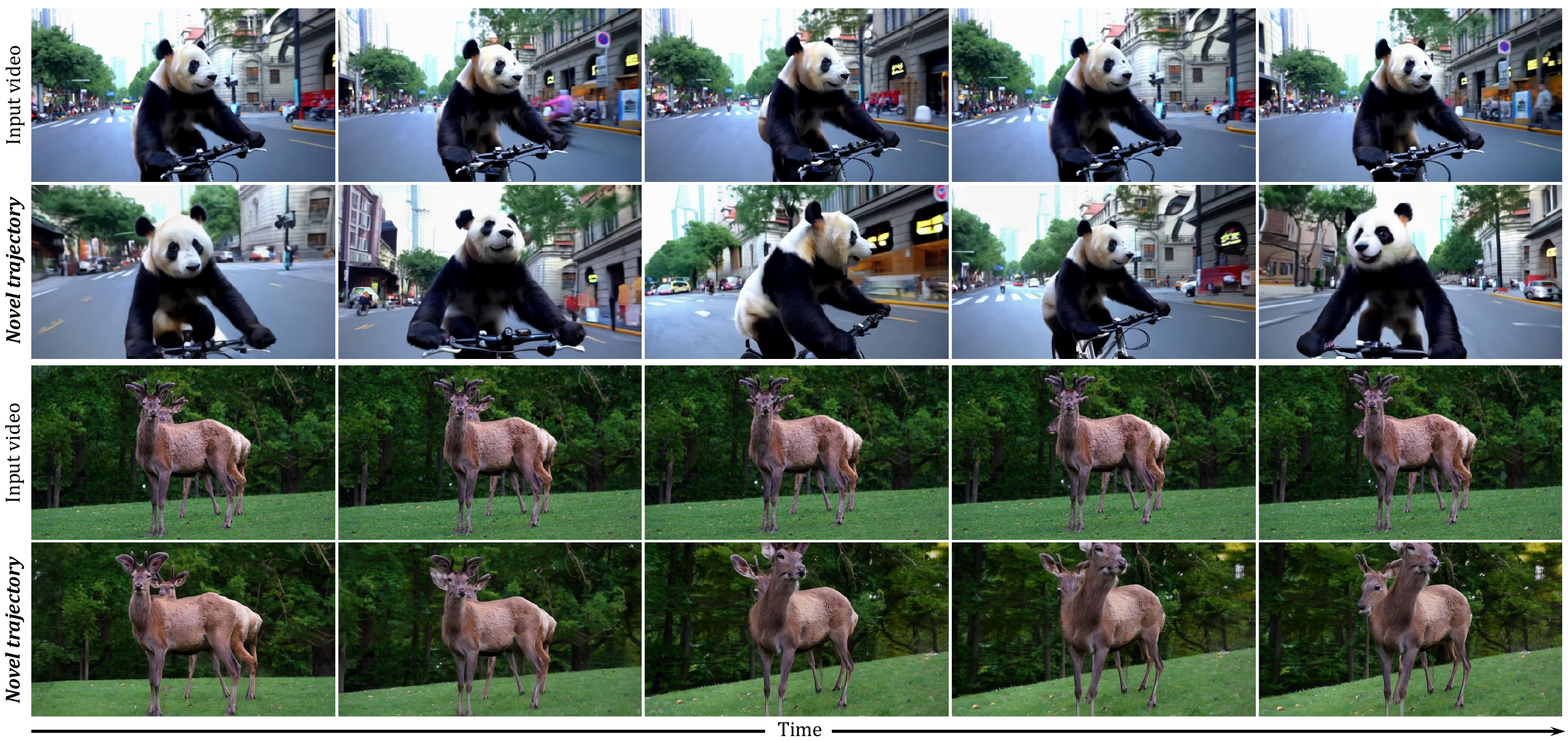}
\vspace{-1.5em}
\captionof{figure}{
    We present \emph{\ourMethod}, a novel approach to redirect camera trajectories for monocular videos, achieving precise control over the view transformations and coherent 4D content generation.
    Please refer to the supplementary \emph{project page} for video results.
    %
    \vspace{1.5em}
    }
\label{fig:teaser}
}]

\footnotetext[2]{~Corresponding author.}

\input{sections/0_abstract}
\input{sections/1_introduction}

\input{sections/2_related}

\input{sections/3_method}

\input{sections/4_experiment}
\input{sections/limitation}
\input{sections/5_conclusion}

\clearpage
{
    \small
    \bibliographystyle{ieeenat_fullname}
    \bibliography{main}
}


\end{document}

%% file: preamble.tex
\usepackage[dvipsnames]{xcolor}


%% file: sections/0_abstract.tex
\begin{abstract}
    We present {\ourMethod}, a novel approach to redirect camera trajectories for monocular videos.
    By disentangling deterministic view transformations from stochastic content generation, our method achieves precise control over user-specified camera trajectories.
    We propose a novel dual-stream conditional video diffusion model that concurrently integrates point cloud renders and source videos as conditions, ensuring accurate view transformations and coherent 4D content generation.
    Instead of leveraging scarce multi-view videos, we curate a hybrid training dataset combining web-scale monocular videos with static multi-view datasets, by our innovative double-reprojection strategy, significantly fostering robust generalization across diverse scenes.
    Extensive evaluations on multi-view and large-scale monocular videos demonstrate the superior performance of our method.
    %

\end{abstract}

%% file: sections/1_introduction.tex
\section{Introduction}
\label{sec:intro}


Videos, whether user-captured or AI-generated, have emerged as a pervasive medium on social media.
Yet, conventional videos merely provide a glimpse into the dynamic world, whereas empowering users to freely redirect the camera trajectory within casual videos promises a more immersive experience.

To this vision, traditional reconstruction-based approaches often yield implausible results in occluded regions~\cite{wang2024shapeom,lee2025fast,liu2023robustdf,stearns2024marbles} or require synchronized multi-view videos~\cite{fridovich2023kplane,li2024spacetime,pumarola2021d,cao2023hexplane}, which remain impractical for most users.
Recent advancements exploit the generative prowess of image and video diffusion models~\cite{rombach2022high,chen2023videocrafter1,blattmann2023svd,blattmann2023align,yang2024cogvideox,lin2024open} to synthesize novel views for static scenes~\cite{yu2024viewcrafter,gao2024cat3d,muller2024multidiff,zeronvs,wu2024reconfusion}, achieving remarkable fidelity from sparse inputs.
However, these methods, tailor-made for static scenes, struggle to deliver plausible videos with novel camera trajectories from dynamic videos.
The most related work is GCD~\cite{gcd}, which adapts a video diffusion model for novel view synthesis from monocular videos with fixed pose, using a synthetic multi-view video training dataset.
Nonetheless, its performance is hindered by the domain gap between synthetic and real videos, and its reliance on implicit pose embeddings~\cite{liu2023zero} limits precise camera trajectory control.

In this paper, we propose \emph{\ourMethod}, a novel framework for generating high-fidelity videos with user-defined camera trajectories from monoocular inputs.
Our approach targets three fundamental challenges:
1). achieving precise trajectory control;
2). ensuring rigorous 4D consistency with the source; and
3). exploiting large-scale, diverse training data for robust generalization.
For trajectory control, we explicitly decouple deterministic view transformation from stochastic content generation, such that we can leverage cutting-edge monocular reconstruction advancements~\cite{hu2024depthcrafter,zhang2024monst3r,lu2024align3r} to accurately model 3D transformations via point cloud renders.
To secure 4D consistency in content generation, we propose a \emph{dual-stream conditional video diffusion model} that integrates both point cloud renders and the source video as conditions.
The point cloud branch utilizes pretrained diffusion layers for precise view transformations, while the source video branch incorporates newly introduced layers to counter occlusions and geometric distortions, thereby elevating the coherence and quality of the generated videos. 
To train our model, synchronized multi-view videos are ideal yet prohibitively scarce.  
Fortunately, our decoupled view transformation and content generation allow us to prepare training data for content generation only.
Motivated by this, We leverage a hybrid training source combining web-scale monocular videos with static multi-view datasets.  
For monocular data, we introduce a novel \emph{double-reprojection strategy} that simulates point cloud renders by reprojecting the source video to a novel view and \emph{back}, aided by video depth estimation.  
For static multi-view data, we employ cutting-edge point cloud reconstruction methods~\cite{mast3r_arxiv24,wang2024dust3r} to derive source and target videos alongside corresponding point cloud renders.  
This innovative data curation approach substantially enriches our training repository, fostering high-fidelity video generation with specfied camera trajectories across diverse unseen scenes.

We extensively evaluate {\ourMethod} on both synchronized multi-view datasets and large-scale monocular video datasets, using reference-based and no-reference video quality assessment metrics, respectively.
Both quantitative and qualitative results demonstrate the superior performance of our method in generating high-fidelity videos with novel camera trajectories, and generalizing robustly across diverse scenes.
Ablation studies also affirm the efficacy of our dual-stream conditioning and dataset curation strategies.
Our contributions are summarized as follows:
\begin{itemize}
    \item We present {\ourMethod}, a novel approach that redirects camera trajectories for monocular videos with exceptional fidelity and broad generalization.
    \item We propose a dual-stream conditioning mechanism that fuses point cloud renders with source videos, ensuring precise trajectory control and coherent 4D generation.
    \item We curate a novel data strategy combining dynamic web-scale monocular video datasets with static multi-view resources, bolstering the model's generalization and robustness across diverse scenes.
\end{itemize}

%% file: sections/2_related.tex
\section{Related Work}
\label{sec:related}

\noindent\textbf{Reconstruction-based novel view synthesis.}
The advent of neural representations like NeRF~\cite{mildenhall2020nerf} and 3DGS~\cite{kerbl20233dgs} has revolutionized static scene novel view synthesis~\cite{barron2021mip, barron2022mip,verbin2022ref,barron2023zip,hu2023Tri-MipRF,liu2024ripnerf,liang2025analytic,zhang2024pixelgs,yu2024mip,muller2022instant,fan2024instantsplat,yu2023nofa,yu2024evagaussians,feng2025ae,li2022nerfacc,chen2021mvsnerf,wang2023sparsenerf,zhu2023fsgs,lin2021barf,garbin2021fastnerf,chen2024mvsplat}, catalyzing significant progress in dynamic novel view synthesis, which is more challenging due to temporal variations.
Current approaches~\cite{hyperreel, fridovich2023kplane, cao2023hexplane, li2024spacetime, pumarola2021d, nerfplayer, yang2023gs4d} predominantly concentrate on reconstructing 4D representations from synchronized multi-view videos, which remain out of reach for typical users.
Early attempts at 4D scene reconstruction from monocular videos employed depth-based warping~\cite{yoon2020nvidia}, later refined by networks to handle occlusions.
Subsequent research~\cite{DynNeRF, sceneflow, nerfies, Tretschk_2021_ICCV,DynIBaR,lee2025fast} utilized neural representations for dynamic scene modeling.
Recent advancements~\cite{stearns2024marbles, wang2024shapeom, gao2024gaussianflow,lei2024mosca} have harnessed the efficiency of 3DGS to generate novel views from monocular videos, often incorporating additional regularizations like optical flow or depth information to enhance performance. 
However, these approaches are limited to reconstructing visible regions, resulting in artifacts when viewed from significantly different perspectives.
%

\input{images_tex/pipeline}

\noindent\textbf{Generative novel view synthesis.}
Diffusion models~\cite{ho2020denoising, song2021denoising, rombach2022high} have demonstrated remarkable promise in generating novel views from monocular images or videos~\cite{zeronvs,huang2024roompainter,zhou2024holodreamer,yu2024hifi,liu2023zero,wang2023motionctrl}.
For static scenarios, Zero-1-to-3~\cite{liu2023zero} presents a camera-pose-conditioned diffusion model for objects with clean backgrounds, later extended to generic scenes by ZeroNVS~\cite{zeronvs}.
ReconFusion~\cite{wu2024reconfusion} refines pose estimation with PixelNeRF~\cite{yu2021pixelnerf} features, whereas CAT3D~\cite{gao2024cat3d} employs ray maps and 3D attention for enhanced consistency.
Recent studies advance novel view synthesis through video diffusion models~\cite{chen2023videocrafter1,blattmann2023svd,xing2023dynamicrafter}, incorporating camera embeddings~\cite{wang2023motionctrl}, depth-warped images~\cite{muller2024multidiff,Ma2024See3D}, Plücker coordinates~\cite{liang2024wonderland,bahmani2024vd3d,xu2024camco,bai2024syncammaster,he2024cameractrl}, and point cloud renderings~\cite{yu2024viewcrafter}.
For dynamic scenes, NVS-Solver~\cite{you2024solver} leverages a pretrained SVD model~\cite{blattmann2023svd} for training-free depth-warped video inpainting but faces challenges with geometric distortions.
CAT4D~\cite{wu2024cat4d} constructs 4D scenes from monocular videos by fine-tuning CAT3D on data produced by diverse video generation methods, risking limited applicability to real-world environments.
Camera-Dolly~\cite{gcd} adapts the SVD model using synthetic multi-view datasets from the Kubric simulator~\cite{greff2022kubric}, yet domain disparities arise with real videos.
DimensionX~\cite{sun2024dimensionx} integrates specialized motion LoRAs for dynamic novel view synthesis but lacks free-view generation.
ReCapture~\cite{zhang2024recapture} re-generates videos with customized camera paths via point cloud rendering and masked fine-tuning, necessitating per-video LoRA adaptation without broader generalization.
Additionally, stereo video generation~\cite{zhao2024stereocrafter} and driving scene novel view synthesis~\cite{fan2024freesim,wang2024freevs,yan2024streetcrafter} are also receiving significant attention.
In contrast, our approach delivers high-fidelity videos with user-defined camera trajectories from generic inputs, circumventing the need for per-video optimization.

%% file: images_tex/pipeline.tex
\begin{figure*}[!t]
  \centering
  \includegraphics[width=1.\textwidth]{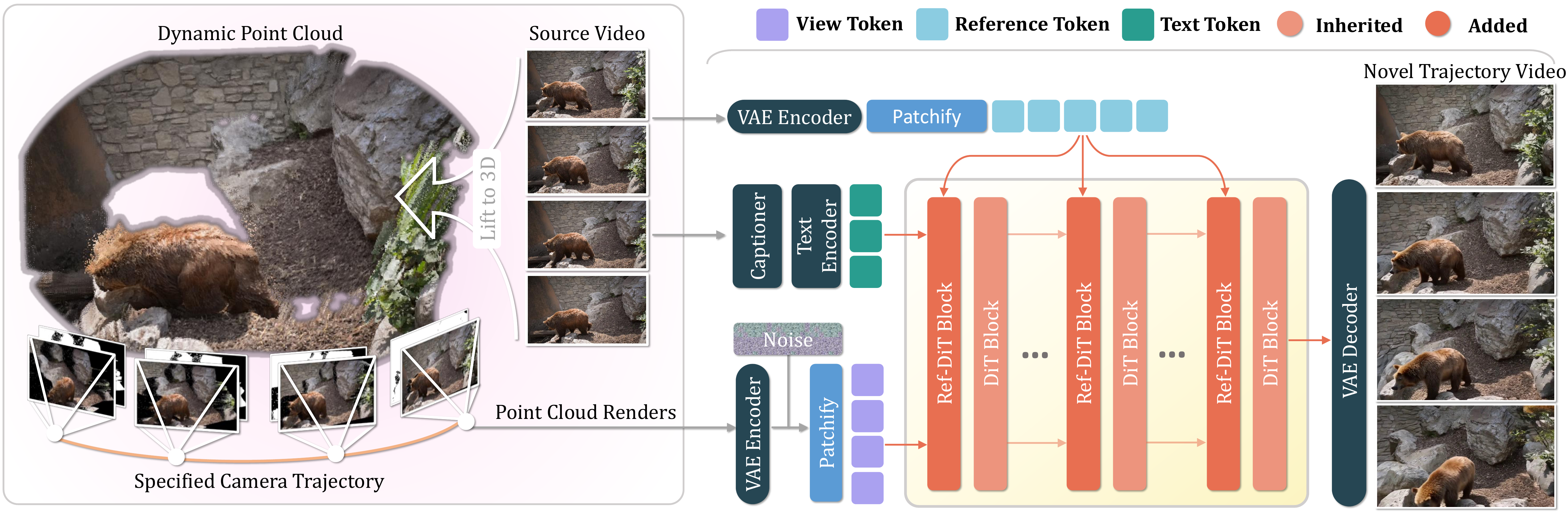}
  \vspace{-1.5em}
  \caption{\textbf{Overview of {\ourMethod}.} 
  Starting with a source video, whether casually captured or AI-generated, we first lift it into a dynamic point cloud via depth estimation. Users can then interactively render the point cloud with desired camera trajectories. Finally, the point cloud renders and the source video are jointly processed by our dual-stream conditional video diffusion model, yielding a high-fidelity video that precisely aligns with the specified trajectory and remains 4D consistent with the source video.
  }
\label{fig:pipeline1}
\end{figure*}

%% file: sections/3_method.tex
\section{Method}
\label{sec:method}

\subsection{Preliminary: Video Diffusion Models}
\label{subsec:prelimiary}
Video diffusion models~\cite{rombach2022high,chen2023videocrafter1,blattmann2023svd,blattmann2023align,yang2024cogvideox,lin2024open} invovle a forward process $q$ to progressively inject noise $\epsilon$ into clean video data $\bm{x}_0 \in \mathbb{R}^{\text{n}\times \text{3}\times \text{h}\times \text{w}}$, creating noisy states $\bm{x}_t = \alpha_t\bm{x}_0 + \sigma_t \epsilon$ over time $t$, 
and a reverse process $p_\theta$ to remove noise via a noise estimator $\epsilon_\theta$, trained by minimizing:
\begin{equation}
\min_{\theta} \mathbb{E}_{t\sim\mathcal{U}(0,1),\epsilon\sim\mathcal{N}(\bm{0},\bm{I})}[\|\epsilon_\theta(\bm{x}_t,t) - \epsilon\|_2^2].
\end{equation}
Following Sora~\cite{sora}, recent diffusion approaches~\cite{lin2024open,yang2024cogvideox} employ the Diffusion Transformer (DiT)~\cite{dit} for the noise estimator. 
During training, a pre-trained 3D VAE encoder compresses videos into latent space $\bm{z}=\mathcal{E}(\bm{x})$. 
Then, $\bm{z}$ is patchified, concatenated with text tokens, and fed into the DiT. 
At inference time, the noise is iteratively denoised into clean tokens, which are mapped back by the VAE decoder to yield the final video $\hat{\bm{x}}=\mathcal{D}(\bm{z})$. 

\subsection{View Transformation via Dynamic Point Cloud}
\label{subsec:pcd} 
Given a source video $\bm{I}^{s} = \{\bm{I}_{\text{i}}^{s}\}_{\text{i=1}}^{\text{n}} \in \mathbb{R}^{\text{n}\times \text{3}\times \text{h}\times \text{w}}$, we aim to explore its underlying 4D scene with a desired camera trajectory.
An overview of our pipeline is illustrated in Fig.~\ref{fig:pipeline1}.
To achieve precise camera trajectory control, we disentangle the deterministic view transformation from the stochastic content generation by lifting the source video into a dynamic point cloud and rendering novel views from it.
%
%
We first estimate a sequence of depth maps $\bm{D}^{s} = \{\bm{D}_{\text{i}}^{s}\}_{\text{i=1}}^{\text{n}} \in \mathbb{R}^{\text{n}\times \text{h}\times \text{w}}$ from the source video via monocular depth estimation~\cite{hu2024depthcrafter,zhang2024monst3r,dptv2}.
Next, we lift the source video into a dynamic point cloud $\bm{P} = \{\bm{P}_{\text{i}}\}_{\text{i=1}}^{\text{n}}$:
\begin{equation} 
    \bm{P}_\text{i} = \Phi^{-1}([\bm{I}^{s}_\text{i}, \bm{D}^{s}_\text{i}], \bm{K}), 
\label{eq:pcd}
\end{equation}
where $\Phi^{-1}$ is the inverse perspective projection, and $\bm{K}  \in \mathbb{R}^{\text{3}\times \text{3}}$ is the camera intrinsic matrix.
With this dynamic point cloud, we render novel views $\bm{I}^{r} = \{\bm{I}_{\text{i}}^{r}\}_{\text{i=1}}^{\text{n}}$ according to a target camera trajectory $\bm{T}^{r} = \{\bm{T}^{r}_{\text{i}}\}_{\text{i=1}}^{\text{n}} \in \mathbb{R}^{\text{n}\times \text{4}\times \text{4}}$:
\begin{equation}
    \bm{I}^{r}_\text{i} = \Phi(\bm{T}^{r}_\text{i}\cdot\bm{P}_i,\bm{K}),
\end{equation}
where $\Phi$ denotes the perspective projection.
%
Because the point cloud is reconstructed in the camera coordinate system, $\bm{T}^{r}$ is defined as the relative transformation from the source camera.
The point cloud renders $\bm{I}^{r}$ have noticeable holes from occlusions and out-of-frame regions, marked by the rendered mask video $\bm{M}^{r} = \{\bm{M}^{r}_{\text{i}}\}_{\text{i=1}}^{\text{n}}$.
Yet, they accurately capture geometric relations and view transformations.
Therefore, we leverage $\bm{I}^{r}$ and $\bm{M}^{r}$ as conditions to guide the generation of high-fidelity videos under the target camera trajectory.

\input{images_tex/cross}

\subsection{Dual-stream Conditional Video Diffusion}
Having the dynamic point cloud renders $\bm{I}^{r}$ and masks $\bm{M}^{r}$, as well as the source video $\bm{I}^{s}$, we learn a conditional distribution $\bm{x} \sim p(\bm{x}~|~\bm{I}^{s},\bm{I}^{r},\bm{M}^{r})$ using a conditional video diffusion model.
We build upon CogVideoX~\cite{yang2024cogvideox,cogfun}, originally designed for text-guided image-to-video (I2V) generation.
We leverage its 3D VAE for video compression/decompression, BLIP~\cite{li2022blip} and T5~\cite{2020t5} encoder for captioning the video, and its DiT blocks for vision-text token processing.

To adapt the I2V model for our task, re-generating videos with desired camera trajectories, we propose a dual-stream conditioning strategy to leverage both the point cloud renders and the source video.
First, we replace the original image conditions with point cloud renders $\bm{I}^{r}$ and masks $\bm{M}^{r}$for precise camera control (Fig.~\ref{fig:pipeline1}).
We encode $\bm{I}^{r}$ and $\bm{M}^{r}$ using the VAE encoder, concatenate them channel-wise with sampled noise, and project them into view tokens.
We then concatenate these view tokens with text tokens along the sequence dimension and feed them into the DiT blocks for denoising.
Since the point cloud renders spatially align with the required view transformations, they provide strong geometric cues to ensure desired camera trajectories.

However, point cloud renders include holes, distortions, and degraded textures due to occlusions and depth errors.
Meanwhile, the source video $\bm{I}^{s}$ offers rich appearance details for high-fidelity generation.
Directly concatenating it is suboptimal, as to be demonstrated in Fig.~\ref{fig:refdit}, due to the spatial misalignment with the target view tokens.
To this end, we encode $\bm{I}^{s}$ into reference tokens via the VAE encoder, and propose a Reference-conditioned Diffusion Transformer (Ref-DiT) block, inserted between the inherited DiT blocks (Fig.~\ref{fig:pipeline1}).
In each Ref-DiT block, as shown in Fig.~\ref{fig:cross}, text and view tokens are processed through 3D attention, followed by a cross-attention layer, where the view tokens act as queries and the reference tokens as keys and values, to inject information from reference tokens into view tokens.
Compared to direct concatenation, our Ref-DiT block enables transferring detailed, yet mis-aligned, information from the source video to the view tokens by the cross-attention mechanism, enhancing the fidelity of the generated novel views.
These two conditioning streams, the point cloud renders and the source video, work synergistically to guide the denoising, enabling generation of high-fidelity videos with accurate camera trajectories and consistent content.

\input{images_tex/warp}

\input{images_tex/comp_nvs}

\subsection{Dataset Curation and Training Scheme}
\label{subsec:data}
To train the model, we ideally require synchronized multi-view videos of diverse 4D scenes.
However, existing multi-view datasets~\cite{corona2021meva, grauman2022ego4d, zheng2023pointodyssey, sener2022assembly101, greff2022kubric} tend to be small in scale, non-photo-realistic in style, or lacking in diversity.
Training solely on such datasets would limit performance and generalization in real-world scenarios.

\noindent\textbf{Data curation.}
Fortunately, our {\ourMethod} explicitly decouples view transformation from content generation, allowing us to only prepare training data for the dual-stream condition video diffusion model.
Motivated by this, we propose two strategies to process web-scale monocular video datasets~\cite{nan2024openvid,wang2024koala,chen2024panda} and static multi-view datasets~\cite{ling2024dl3dv,zhou2018real10k}, respectively.

For monocular datasets, we propose a {double-reprojection} strategy to produce large-scale training pairs of point cloud renders and target videos, serving as condition and supervision.
Given a target video $\bm{I}^{t}$, we lift it into a dynamic point cloud $\bm{P}^{t}$ using Eq.~\ref{eq:pcd}, then render novel views $\bm{I}^{\prime}$ and corresponding depths $\bm{D}^{\prime}$ with a randomly sampled relative view transformation $\Delta\bm{T}$.
%
Next, we back-project $\bm{I}^{\prime}$ into point cloud $\bm{P}^{\prime}$ with $\bm{D}^{\prime}$, and re-render the view $\bm{I}^{\prime\prime}$ with the inverse transformation $\Delta\bm{T}^{-1}$.
%
As shown in Fig.~\ref{fig:warp}, $\bm{I}^{\prime}$ has a view transformation from target video $\bm{I}^{t}$, while $\bm{I}^{\prime\prime}$ realigns with $\bm{I}^{t}$ yet contains holes from disocclusions and out-of-frame regions, like the point cloud renders.
In this way, we can generate large-scale, diverse, and high-quality training pairs from web-scale monocular video datasets.
%
We adopt the OpenVid-1M~\cite{nan2024openvid} monocular video dataset and employ DepthCrafter~\cite{hu2024depthcrafter} to create $60 \text{K}$ paired point cloud renders and target videos for training.

Another valuable data source is static multi-view datasets~\cite{ling2024dl3dv,zhou2018real10k}, which feature diverse and large camera movements.
To incorporate these datasets, we design a paradigm to generate large-scale training triplets, consisting of a source video, a target video, and a point cloud render.
For each video in the dataset, we sample a segment and simuteously reconstruct its global point cloud and estimate camera pose for each frame, using advanced reconstruction method, \eg MASt3R~\cite{mast3r_arxiv24}.
We then sample two video clips from the segment, with overlapping content, designating one as the source video and the other as the target video.
Finally, the point cloud of the source video is rendered to match the camera pose of the target video, creating the required triplet for training. 
We adopt the DL3DV~\cite{ling2024dl3dv} and RealEstate10K~\cite{zhou2018real10k} datasets, generating $120 \text{K}$ static multi-view training data.

\noindent\textbf{Training scheme.}
Having the datasets, we adopt a two-stage training scheme to leverage both dynamic monocular and static multi-view data.
In the first stage, we train the DiT blocks and the 3D attention layers in the Ref-DiT blocks using all data, but freeze the cross-attention layers and skip reference-token injection.
This step focuses on synthesizing missing regions and correcting geometric distortions, while preserving the targeted view transformations in the point cloud renders.
However, without reference tokens, the generated content may be inconsistent with the source video in 4D space.
In the second stage, we enable the model to transfer detailed information from the reference tokens to the view tokens, by training on the triplet training data.
Since only the multi-view static data provides the required triplets, we use it exclusively in the second stage.
To prevent overfitting to static data, we only train the cross-attention and patch embedding layers in the Ref-DiT blocks, while freezing all other modules.
After these two stages, the model is capable of generating high-fidelity videos with accurate view transformations and consistent content.

%% file: images_tex/cross.tex
\begin{figure}[!t]
  \centering
  \includegraphics[width=.93\linewidth]{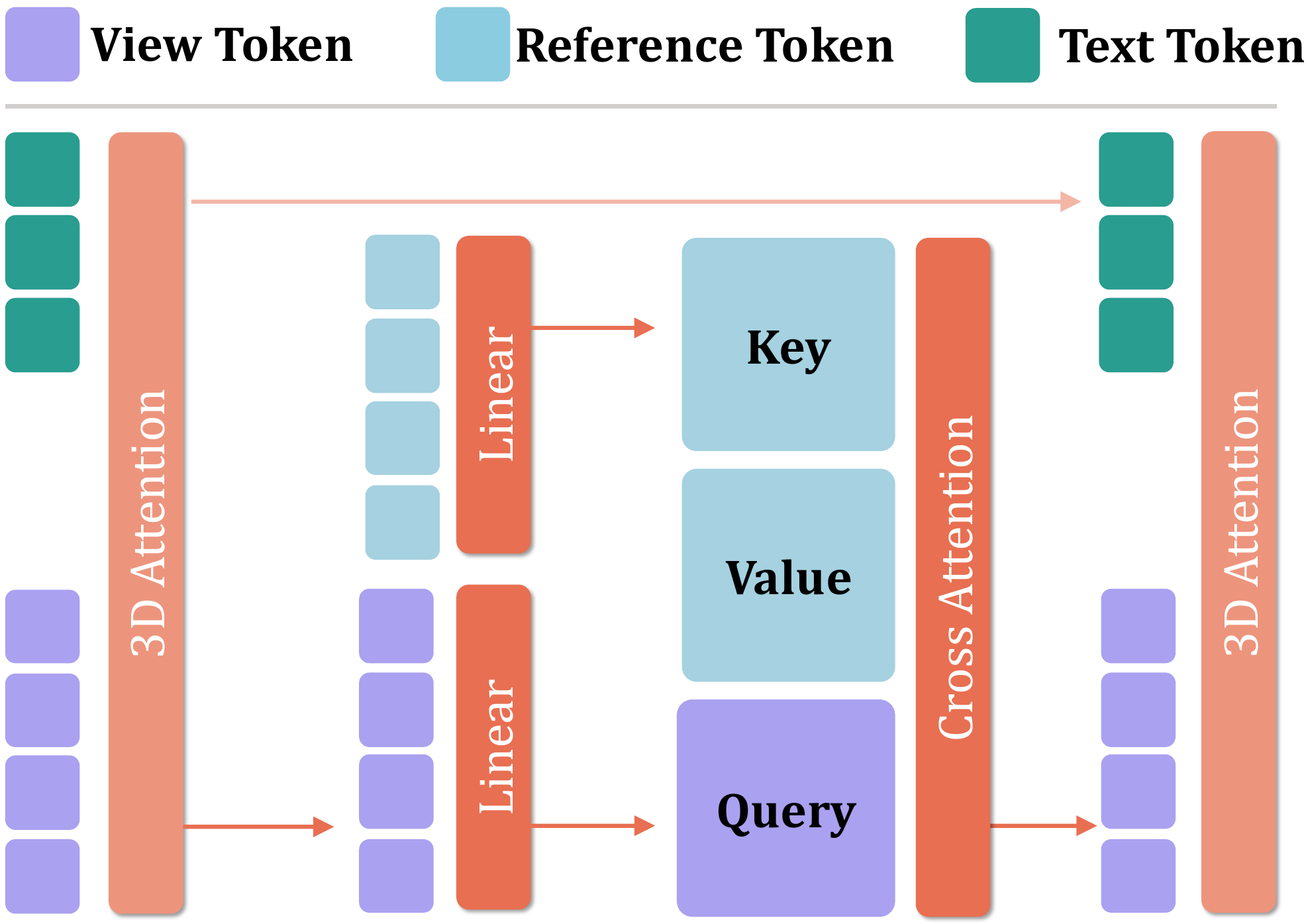}
\caption{\textbf{Ref-DiT Block.} The text and view tokens are first processed through 3D attention, followed by a cross-attention that injects the detailed, yet mis-aligned, reference information into the view tokens, yielding refined view tokens for subsequent layers.}
  \vspace{-0.5em}
\label{fig:cross}
\end{figure}

%% file: images_tex/warp.tex
\begin{figure}[t]
  \centering
  \includegraphics[width=1.\linewidth]{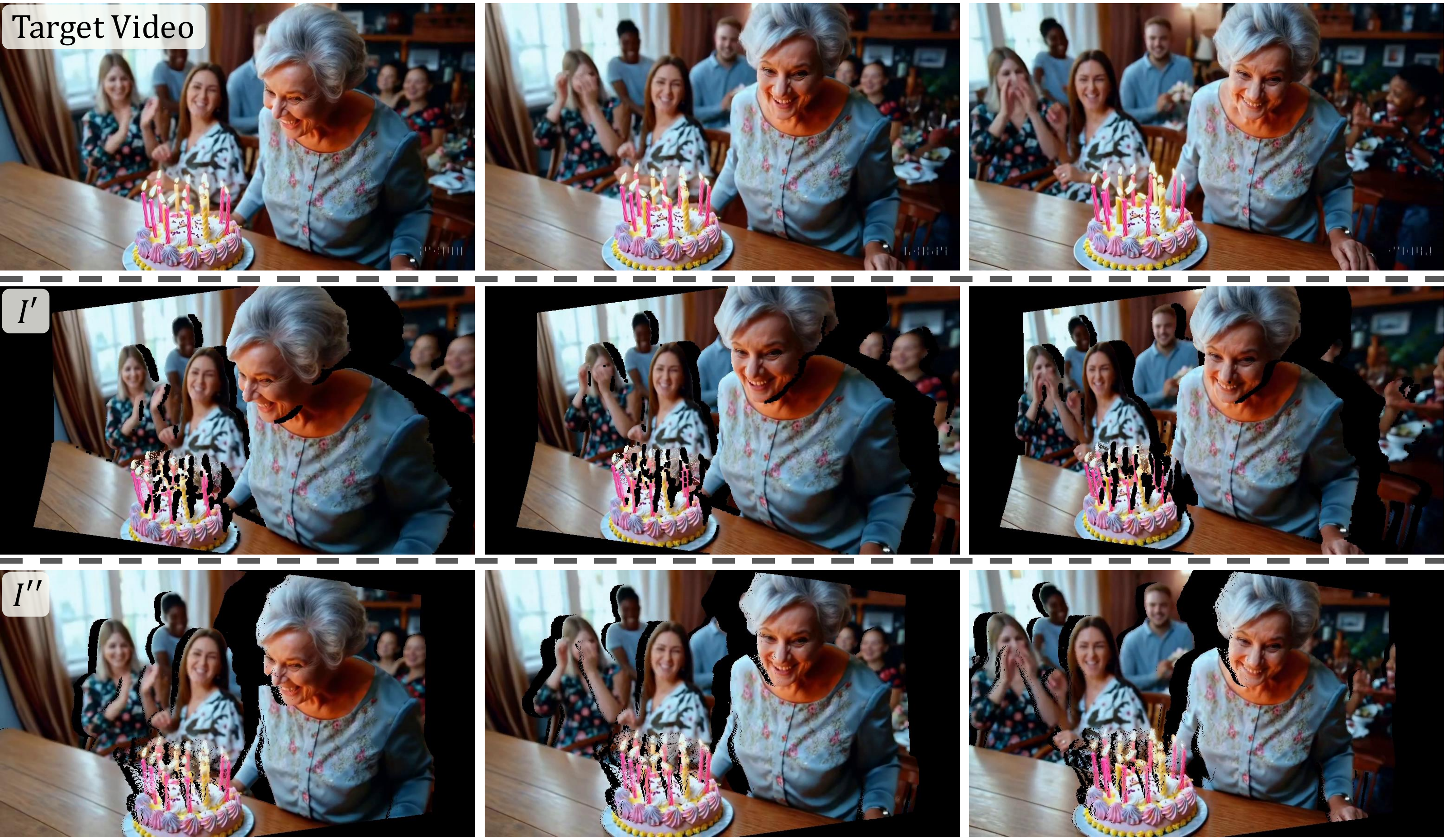}
  \vspace{-1.5em}
  \caption{\textbf{Double-reprojection.} Given a target video, we lift it into a dynamic point cloud to render a novel view $\bm{I}^{\prime}$ via a random view transformation. Then $\bm{I}^{\prime}$ is reprojected to the original camera pose, yielding $\bm{I}^{\prime\prime}$ through the inverse view transformation. $\bm{I}^{\prime\prime}$ contains occlusions and aligns with the target video, simulating the point cloud renders.}
  \vspace{-0.5em}
\label{fig:warp}
\end{figure}

%% file: images_tex/comp_nvs.tex
\begin{figure*}[t]
  \centering
  \includegraphics[width=1.\textwidth]{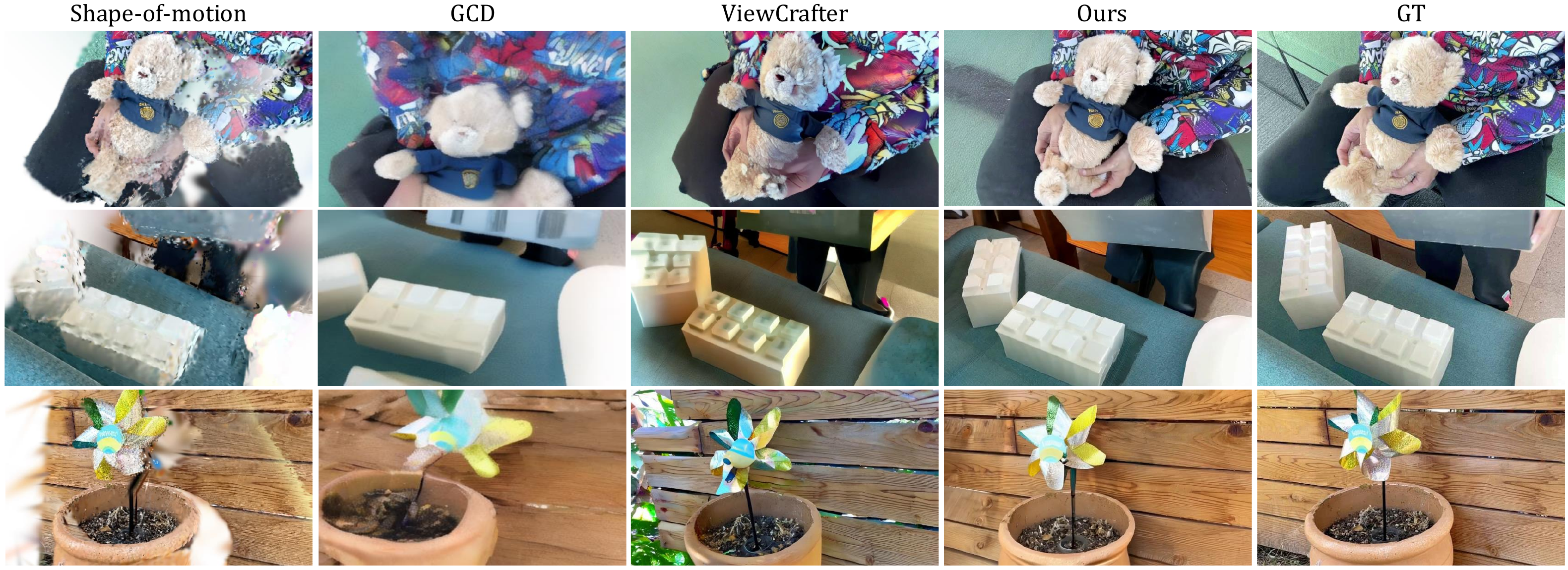}
  \vspace{-1.5em}
  \caption{\textbf{Qualitative comparison of novel trajectory video synthesis.}
  We compare our method with both reconstruction-based method, Shape-of-motion~\cite{wang2024shapeom}, and generative methods, GCD~\cite{gcd} and ViewCrafter~\cite{yu2024viewcrafter} on the multi-view dataset, iphone~\cite{iphone}.
  }
  \vspace{-1em}
\label{fig:compare_nvs}
\end{figure*}

%% file: sections/4_experiment.tex
\section{Experiments}
\label{sec:experiments}

\subsection{Implementation}
\label{subsec:implementation}
We implement our dual-stream conditional video diffusion model based on the pretrained CogVideoX-Fun-5B~\cite{cogfun,yang2024cogvideox} architecture.
During training, the frame resolution is fixed at 384$\times$672, and the video length is set to 49 frames.
The first training stage is conducted for 10,000 iterations with a learning rate of 1$\times$10$^{-5}$, while the second stage is conducted for 5,000 iterations with a learning rate of 2$\times$10$^{-6}$.
Both training stages use a mini-batch size of 8 and are conducted on eight GPUs.
For producing dynamic point cloud, we use DepthCrafter~\cite{hu2024depthcrafter} to estimate temporally consistent depth sequences from the source video and empirically set camera intrinsic parameters.
%

\input{tables/table_nvs}
\input{tables/table_vbench}

\subsection{Evaluation on Multi-view Video Benchmark}
\label{subsec:comp_4d}


\noindent\textbf{Dataset and evaluation metrics.}
To pixel-wisely evaluate the quality of the synthesized videos under novel trajectories, we employ the iPhone dataset~\cite{iphone}, which contains 7 scenes captured with a casually moving camera and two static cameras.
We take the casually-captured video as the source and the first fixed-camera video as the target for evaluation, since the second target video is incomplete.
Following~\cite{wang2024shapeom}, we discard the ``Space-out'' and ``Wheel'' scenes due to camera and LiDAR errors, and use the remaining 5 refined scenes with COLMAP poses and aligned depth for evaluation.
%
We adopt PSNR, SSIM, and LPIPS~\cite{zhang2018unreasonable} as evaluation metrics.


\noindent\textbf{Comparison baselines.}
We compare our method with three baselines: GCD~\cite{gcd}, ViewCrafter~\cite{yu2024viewcrafter}, and Shape-of-motion~\cite{wang2024shapeom}, where the first two methods are diffusion-based generative novel view synthesis models, and the last one is the SOTA reconstruction-based 4D novel view synthesis method.
GCD is a 4D novel view synthesis method that incorporates implicit camera pose embeddings into a video diffusion model, and can generate 14-frame novel view videos from a source video with given relative camera pose changes.
ViewCrafter is originally designed for static novel view synthesis from single or sparse images through the point cloud renders, but we found that it can be also used for 4D novel view synthesis by conditioning it on 4D point cloud renders.
%
%
Shape-of-motion leverages dense point tracks, monocular depths, and motion masks as regularization to optimize a dynamic 3DGS from monocular video.


\noindent\textbf{Qualitative results.}
We first compare the visual quality of results by our method and the baselines in Fig.~\ref{fig:compare_nvs}, where the ground truth novel trajectory videos are shown in the last column.
We can see that the videos synthesized by Shape-of-motion exhibit significant holes due to its reconstruction-based nature.
And the results of GCD suffer from overly smoothed details and view misalignment.
This is because GCD is trained on synthetic data with poor texture quality that significantly differs from real-world videos, and its view control mechanism is through an implicit pose embedding, which lacks precise control.
Results from ViewCrafter show better pose accuracy but struggle to generate high-fidelity frames since it is designed for static novel view synthesis and lacks the ability to maintain 4D coherence.
In contrast, our method generates novel trajectory videos with high fidelity, 4D consistency, and precise pose control, thanks to the dual-stream conditional video diffusion model and the diverse web-scale training data.

\noindent\textbf{Quantitative comparison.}
We further quantitatively evaluate the results of our method and the baselines in Tab.~\ref{tab:nvs}.
The results show that our method consistently outperforms the baselines across all metrics by a significant margin, indicating that our approach maintains better quality of the synthesized novel trajectory videos and a closer resemblance to the ground truth.


\input{images_tex/comp_vbench.tex}

\subsection{Evaluation on in-the-wild Video Benchmark}
\label{subsec:vbench}
\noindent\textbf{Dataset and evaluation metrics.}
The previous multi-view video benchmark contains a limited number of scenes, making it hard to evaluate the generalization ability.
Moreover, using pixel-wise metrics like PSNR and SSIM to access similarity between generated and ground truth novel views is not ideal, as occluded regions can be filled with many semantically plausible contents, yet not resemble the ground truth.
This is also reflected by the values in Tab.~\ref{tab:nvs}, where the PSNR values for all methods are relatively low (lower than 15 dB).
For better evaluation, we collect a relatively large-scale in-the-wild video benchmark, comprising \textbf{100} real-world monocular videos~\cite{davis} and \textbf{60} high-quality videos generated by T2V models~\cite{sora,vidu,kling,hunyuan,pika}.
For each video, we generate 6 different novel trajectory videos by our method and the generative baselines, GCD and ViewCrafter.
For evaluation metrics, we use the VBench~\cite{vbench} protocol, which includes Subject Consistency, Background Consistency, Temporal Flickering, Motion Smoothness, Aesthetic Quality, Imaging Quality, and Overall Consistency scores.

\input{images_tex/refdit.tex}


\noindent\textbf{Quantitative results.}
%
As shown in Tab.~\ref{tab:vbench}, our method significantly outperforms the baselines across all video quality metrics, demonstrating state-of-the-art performance in generating high-quality novel trajectory videos, in terms of subject and background consistency, temporal flickering, motion smoothness, aesthetic quality, and imaging quality.

\noindent\textbf{Qualitative results.}
Fig.~\ref{fig:compare_vbench} presents an example of novel trajectory video generation, illustrating the target camera trajectory as \emph{zoom-in and orbit to the right}.
Results from GCD exhibit noticeable artifacts, oversmoothed textures, and pronounced inaccuracies in camera motion, resulting in misalignment with the intended trajectory.
While ViewCrafter~\cite{wang2023motionctrl} demonstrates improved pose accuracy, it struggles to preserve consistency with the source video, leading to content drift and color discrepancies.
By contrast, our method achieves higher visual fidelity, superior alignment with the target trajectory, and consistent content with the source video.


\input{tables/table_ablation}

\subsection{Ablation Study}
\label{subsec:ablation}

\noindent\textbf{Effectiveness of the Ref-DiT blocks.}
In one of our key contributions, dual-stream conditional video diffusion model, we propose the Ref-DiT blocks to incorporate the source video information into the denoising process.
To evaluate the effectiveness of the Ref-DiT blocks, we compare the novel view synthesis results of three variants: 
\textbf{w/o Ref-DiT}: the base model without Ref-DiT blocks,
\textbf{w/ Concat Condition}: directly concatenating the source video with the point cloud renders, and
\textbf{w/ Ref-DiT}: our full model.
%
%
From the qualitative results in Fig.~\ref{fig:refdit}, we can observe that the model without Ref-DiT blocks struggles to maintain content consistency, \eg the eyebrows marked in the yellow box.
This is because the point cloud renders may suffer from geometric distortions and texture artifacts, and the diffusion model cannot effectively generate content consistent with the source video.
And the simple concatenation strategy also fails to generate plausible results, since the source video information is spatially misaligned with the target view and direct concatenation cannot effectively integrate the source video information.
In contrast, our full model generates videos with high fidelity and content consistency, demonstrating the effectiveness of the Ref-DiT blocks.
Furthermore, we quantitatively ablate the effectiveness of the Ref-DiT blocks on the iPhone dataset~\cite{iphone} in Tab.~\ref{tab:ablation}.
Our full model significantly outperforms the other two variants across all metrics, which confirms that the Ref-DiT blocks significantly enhance the quality of novel trajectory video generation.

\input{images_tex/mix}

\noindent\textbf{Ablation on the training data.}
The dataset curation strategy is another key contribution of our work, where we train our model on a mixed dataset of both dynamic monocular and static multi-view data.
To validate its effectiveness, we conduct an ablation study to compare the novel trajectory video generation results of models trained without dynamic data, without multi-view data, and with the mixed data.
The qualitative results are presented in Fig.~\ref{fig:mix}, where we can observe that the model trained without multi-view data struggles to address occlusions and geometric distortions effectively, \eg the yellow box marked regions.
Conversely, the model trained without dynamic data fails to accurately follow the motion of the source video.
In contrast, training with the mixed data allows the model to generate semantically meaningful content in the occlusion regions and maintain consistent motion with the source video.
The quantitative results in Tab.~\ref{tab:ablation} further confirm the effectiveness of our dataset curation and training strategy.

\input{images_tex/limit}

%% file: tables/table_nvs.tex
\begin{table*}[]
    \setlength{\tabcolsep}{2.9pt}
\centering
\small
\caption{\textbf{Quantitative comparison of novel trajectory video synthesis.} 
We report the PSNR, SSIM, and LPIPS metrics for each scene and the mean values across all scenes on the multi-view dataset, iphone~\cite{iphone}. The best results are highlighted in \textbf{bold}.}
\vspace{-0.7em}
\resizebox{\textwidth}{!}{
\begin{tabular}{l || c c c c c c | c c c c c c | c c c c c c }
\toprule
& \multicolumn{6}{c}{PSNR $\uparrow$} & \multicolumn{6}{c}{SSIM $\uparrow$} & \multicolumn{6}{c}{LPIPS $\downarrow$} \\

Method & \textit{Apple} & \textit{Block} & \textit{Paper} & \textit{Spin} & \textit{Teddy} & \textbf{Mean} & \textit{Apple} & \textit{Block} & \textit{Paper} & \textit{Spin} & \textit{Teddy} & \textbf{Mean} & \textit{Apple} & \textit{Block} & \textit{Paper} & \textit{Spin} & \textit{Teddy} & \textbf{Mean} \\

\hline
GCD & 9.823 & 12.30 & 9.75 & 10.37 & 11.61 & 10.77 & 0.215 & 0.458 & 0.398 & 0.324 & 0.385 & 0.356 & 0.738 & 0.590 & 0.535 & 0.576 & 0.629 & 0.614 \\

ViewCrafter & 10.19 & 10.28 & 10.63 & 11.15 & 11.50 & 10.75 & 0.245 & 0.427 & 0.344 & 0.308 & 0.372 & 0.339 & 0.750 & 0.615 & 0.521 & 0.533 & 0.606 & 0.605 \\

Shape-of-motion & 11.06 & 11.72 & 11.93 & 11.28 & 10.42 & 11.28 & 0.197 & 0.446 & 0.425 & 0.319 & 0.357 & 0.349 & 0.879 & 0.601 & 0.486 & 0.560 & 0.650 & 0.635 \\

\cellcolor{light}Ours & \cellcolor{light}\textbf{13.88} & \cellcolor{light}\textbf{14.21} & \cellcolor{light}\textbf{14.89} & \cellcolor{light}\textbf{14.51} & \cellcolor{light}\textbf{13.73} & \cellcolor{light}\textbf{14.24} & \cellcolor{light}\textbf{0.285} & \cellcolor{light}\textbf{0.528} & \cellcolor{light}\textbf{0.482} & \cellcolor{light}\textbf{0.380} & \cellcolor{light}\textbf{0.411} & \cellcolor{light}\textbf{0.417} & \cellcolor{light}\textbf{0.612} & \cellcolor{light}\textbf{0.479} & \cellcolor{light}\textbf{0.471} & \cellcolor{light}\textbf{0.518} & \cellcolor{light}\textbf{0.513} & \cellcolor{light}\textbf{0.519} \\

\bottomrule
\end{tabular}
}
\label{tab:nvs}
\end{table*}

%% file: tables/table_vbench.tex
\begin{table*}[]
\setlength{\tabcolsep}{3pt}
\centering
\small

\caption{
    \textbf{VBench results on in-the-wild monocular videos.}
    We compiled a large-scale in-the-wild video benchmark with 100 real-world and 60 high-quality T2V-generated videos, and report the VBench scores of novel trajectory videos from GCD~\cite{gcd}, ViewCrafter~\cite{yu2024viewcrafter}, and our method.
    The best results are highlighted in \textbf{bold}.
    }
    \vspace{-0.7em}
\resizebox{\textwidth}{!}{%
\begin{tabular}{l || c  c c c c c c }
\toprule
& \multicolumn{7}{c}{VBench $\uparrow$} \\

Method & Subject Consis. & Background Consis. & Temporal Flicker. & Motion Smooth. & Overall Consis. & Aesthetic Quality & Imaging Quality \\

\hline
GCD & 0.7677 & 0.8533 & 0.8215 & 0.8950 & 0.2321 & 0.4154 & 0.5203 \\

ViewCrafter & 0.7305 & 0.8782 & 0.8954 & 0.9031 & 0.2432 & 0.4950 & 0.5548 \\

\cellcolor{light}Ours & \cellcolor{light}\textbf{0.9236} & \cellcolor{light}\textbf{0.9512} & \cellcolor{light}\textbf{0.9437} & \cellcolor{light}\textbf{0.9815} & \cellcolor{light}\textbf{0.2847} & \cellcolor{light}\textbf{0.5920} & \cellcolor{light}\textbf{0.6479} \\

\bottomrule
\end{tabular}%
}
\label{tab:vbench}
\vspace{-1em}
\end{table*}

%% file: images_tex/comp_vbench.tex
\begin{figure*}[!t]
  \centering
  \includegraphics[width=1.0\textwidth]{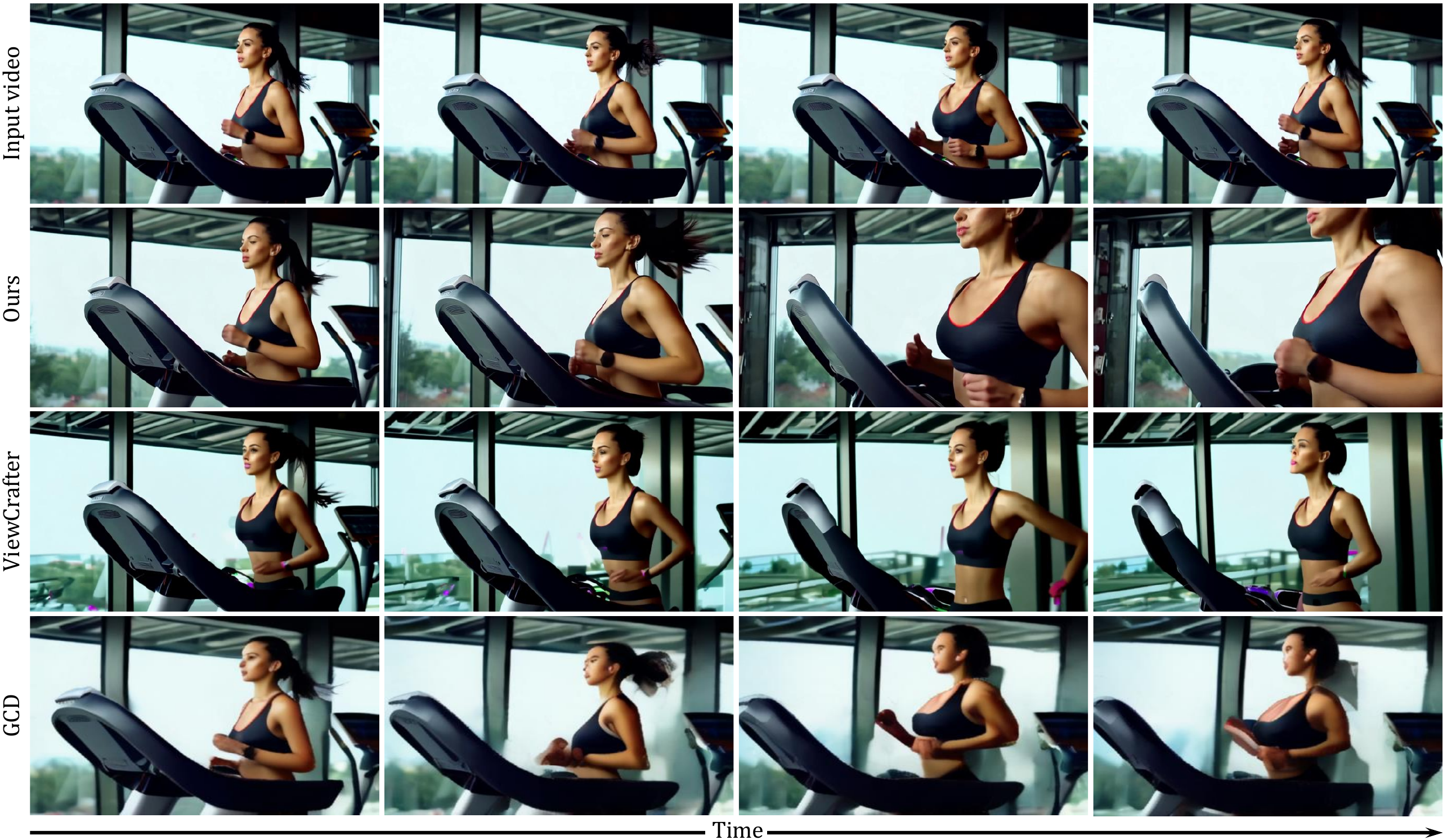}
  \vspace{-1.5em}
  \caption{
    \textbf{Qualitative comparison on in-the-wild monocular videos.}
    We show results of redirecting the camera trajectory as \emph{``zoom-in and orbit to the right''} from the input videos, produced by our method and the generative baselines, GCD~\cite{gcd} and ViewCrafter~\cite{yu2024viewcrafter}.
  }
\label{fig:compare_vbench}
\vspace{-1em}
\end{figure*}

%% file: images_tex/refdit.tex
\begin{figure}[t]
  \centering
  \includegraphics[width=1.\linewidth]{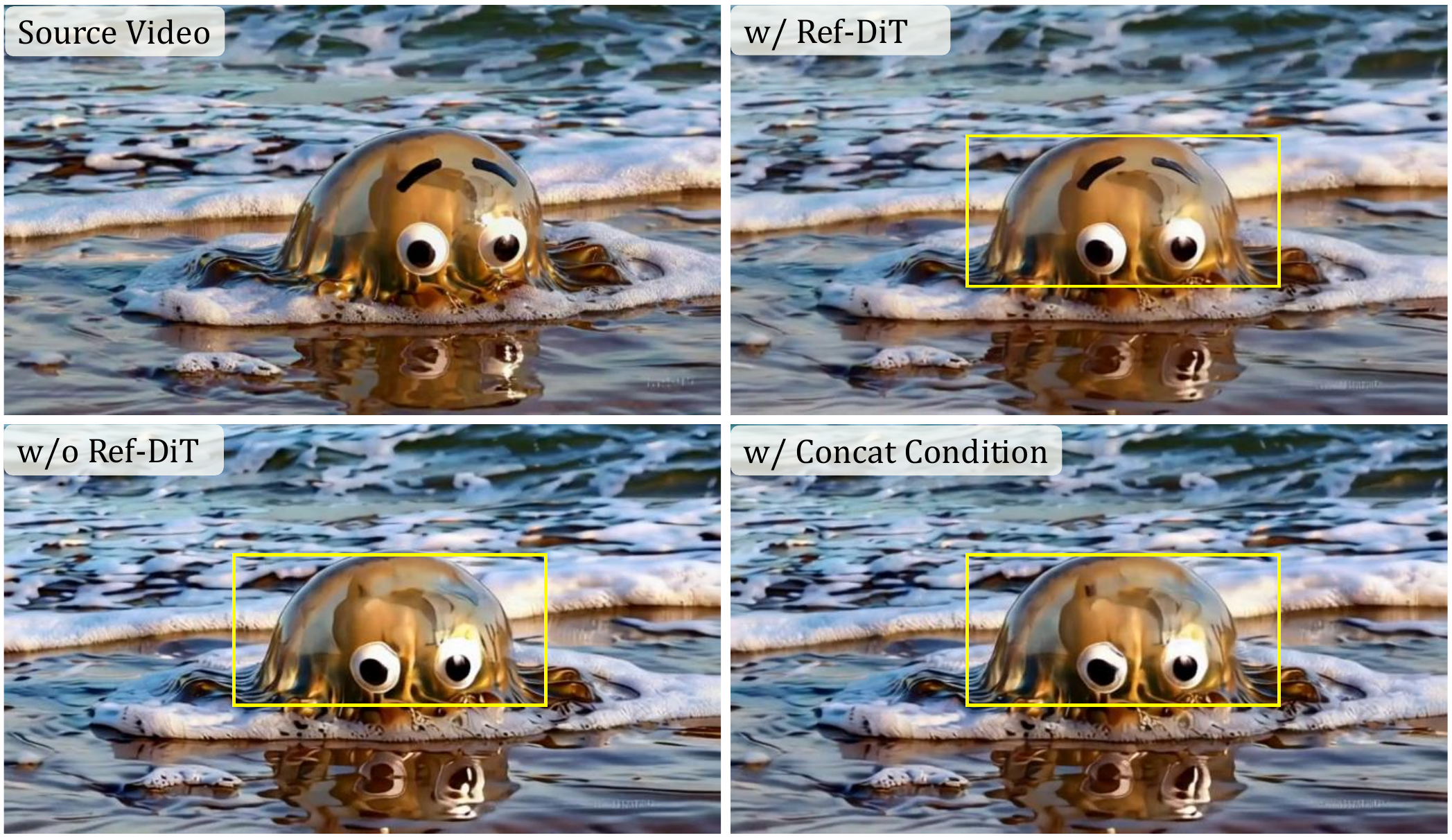}
  \vspace{-1.5em}
  \caption{
\textbf{Effectiveness of Ref-DiT blocks.} 
We compare our full model (w/ Ref-DiT) to two alternatives: a baseline without Ref-DiT (w/o Ref-DiT), and a variant that directly concatenates the source video with the point cloud renders (w/ Concat Condition).
The yellow box highlights the most prominent differences.
}
  \vspace{-2em}
\label{fig:refdit}
\end{figure}

%% file: tables/table_ablation.tex
\begin{table}[tbp]
\centering
\small

\caption{\textbf{Ablation study.} 
We report the PSNR, SSIM, and LPIPS metrics for the full model and its ablated versions on the multi-view dataset, iphone~\cite{iphone}.
The best results are highlighted in \textbf{bold}.}
\vspace{-0.7em}
\resizebox{\linewidth}{!}{%
\begin{tabular}{l || c c c}
\toprule
Method & {PSNR $\uparrow$} & {SSIM $\uparrow$} & {LPIPS $\downarrow$} \\
\hline
Ours w/o Ref-DiT & 11.63 & 0.341 & 0.612  \\

Ours w/ concat. condition & 11.05 & 0.352 & 0.627 \\

Ours w/o dynamic data & 10.93 & 0.348 & 0.621  \\

Ours w/o mv data & 12.35 & 0.361 & 0.588  \\

\cellcolor{light}Ours full & \cellcolor{light}\textbf{14.24} & \cellcolor{light}\textbf{0.417} & \cellcolor{light}\textbf{0.519}  \\

\bottomrule
\end{tabular}%
}
\vspace{-1em}

\label{tab:ablation}
\end{table}

%% file: images_tex/mix.tex
\begin{figure}[tbp]
  \centering
  \includegraphics[width=1.\linewidth]{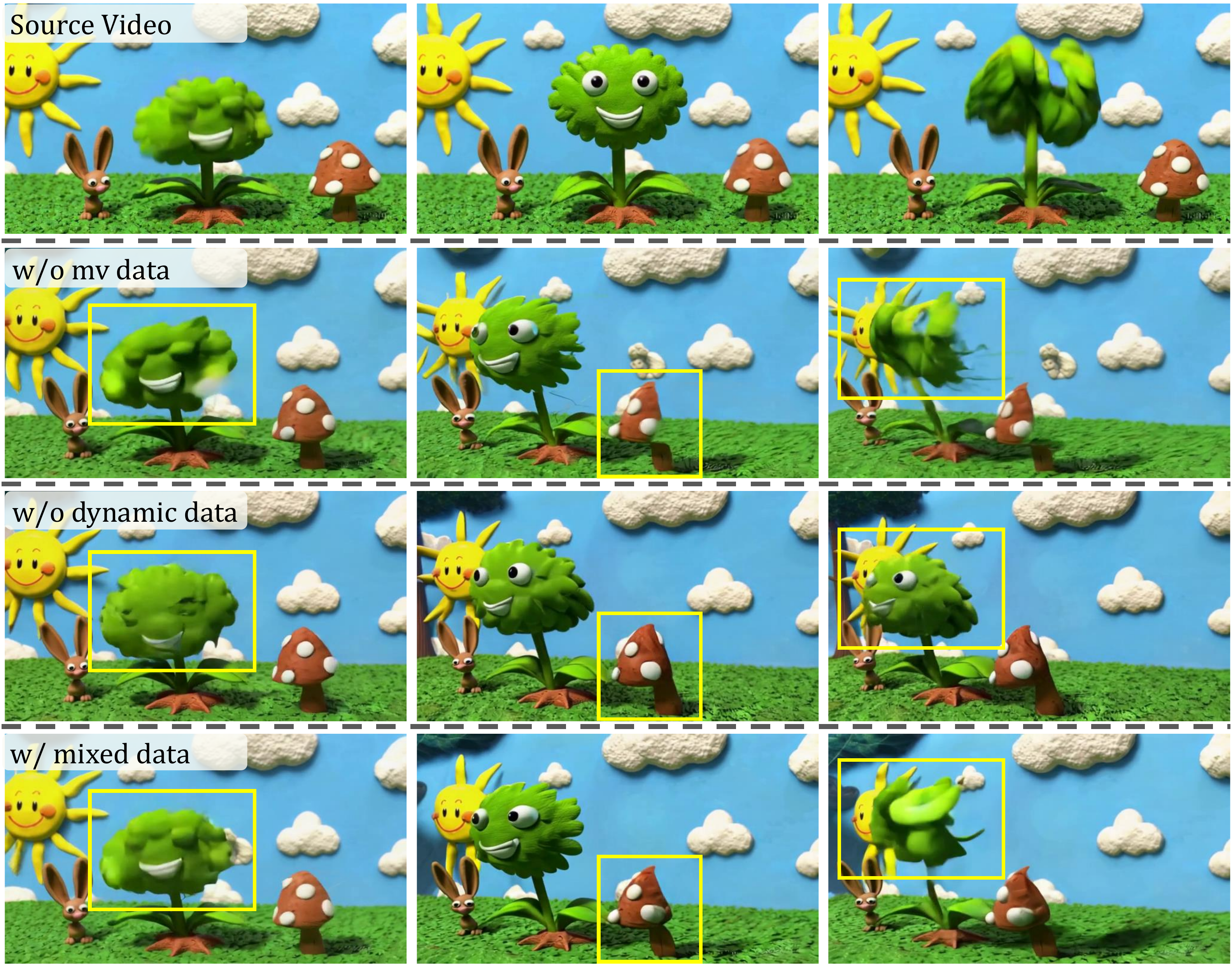}
  \vspace{-1.5em}
  \caption{
    \textbf{Ablation on the training data.} 
    We compare our model trained with mixed data to two alternatives: training without multi-view data and training without dynamic data. 
    The yellow box highlights the most prominent differences of occulusions, geometric distortions, and motion consistency.
    }
  \vspace{-1.8em}
\label{fig:mix}
\end{figure}

%% file: images_tex/limit.tex
\begin{figure}[tbp]
  \centering
  \includegraphics[width=1.\linewidth]{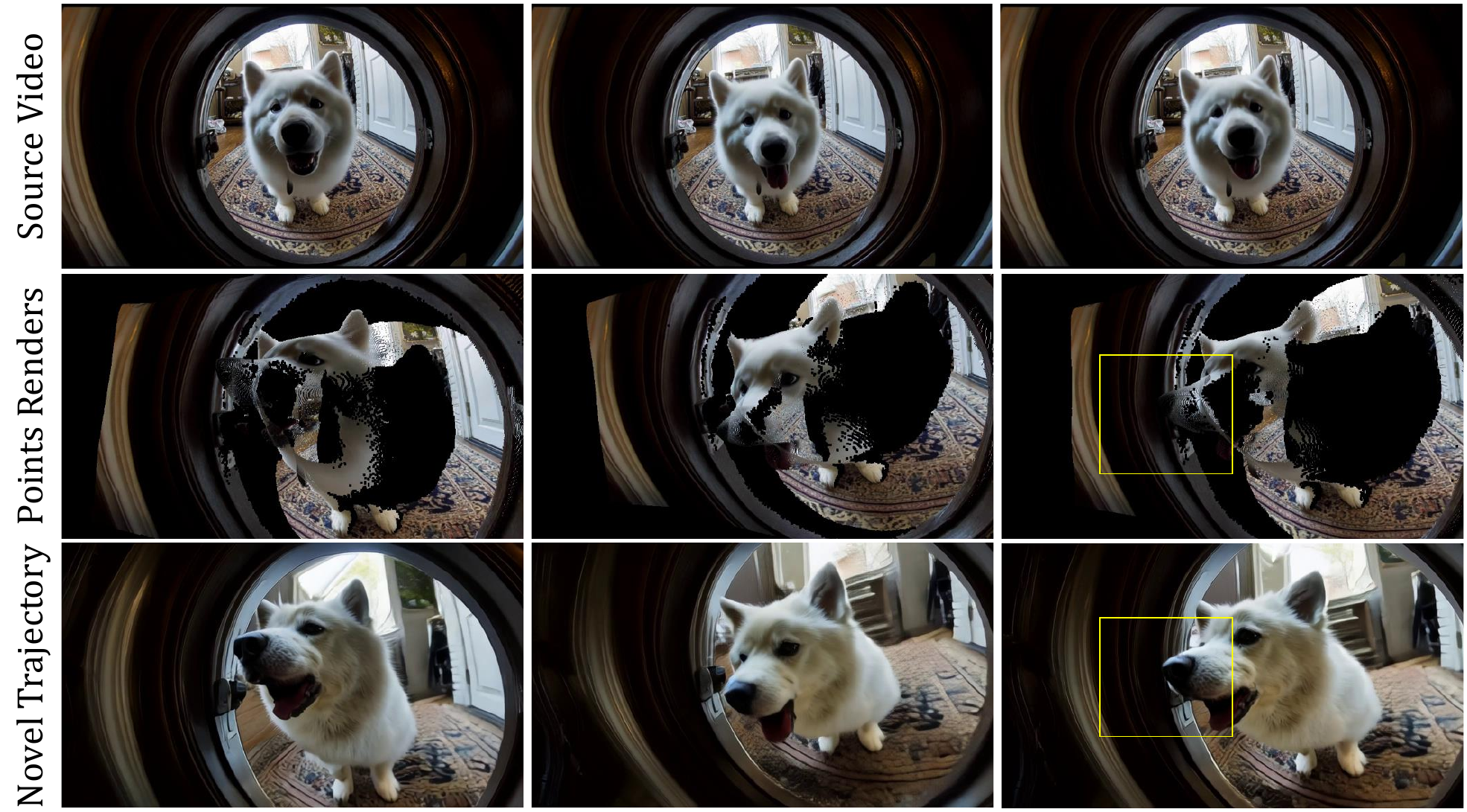}
  \vspace{-1.5em}
  \caption{
    \textbf{Failure case.} 
    Due to inaccuracies in depth estimation, the generated novel trajectory videos may exhibit physically implausible behavior, \eg, the dog's nose appears to pass through the glass of the door.
    }
  \vspace{-1.5em}
\label{fig:limit}
\end{figure}

%% file: sections/limitation.tex
\subsection{Limitations}
\vspace{-0.5em}
While our method can generate high-fidelity novel trajectory videos, it still has several limitations.
Firstly, it struggles to synthesize very large-range trajectories, such as 360-degree viewpoints, due to insufficient 3D cues from monocular inputs and the constrained generation length of the video diffusion model.
Secondly, because our approach relies on depth estimation to create dynamic point clouds from monocular videos, inaccuracies in this process may propagate, producing suboptimal novel trajectory videos (see Fig.~\ref{fig:limit}).
Finally, as a video diffusion model, our method involves multi-step denoising during inference, leading to relatively high computational overhead.

%% file: sections/5_conclusion.tex
\section{Conclusion}

We introduce {\ourMethod}, an innovative approach to redirect camera trajectories for monocular videos.
It allows users to re-generate high-fidelity videos with desired camera trajectories from casual captures or AI-generated footage, while maintaining 4D consistency with the source video.
Our dual-stream conditional video diffusion model integrates point cloud renders and source videos as condition, ensuring accurate view transformations and coherent 4D content generation.
Instead of relying on scarce synchronized multi-view video datasets, we assemble a hybrid training corpus from large-scale dynamic monocular videos and static multi-view datasets, significantly enhancing model generalization across diverse scenarios.
Extensive evaluations on multi-view and large-scale monocular videos confirm our method’s performance in producing high-fidelity outputs with novel camera trajectories.